\documentclass[twoside, 12pt]{report}
\usepackage[english]{babel}
\usepackage{epsfig}
\usepackage{psfig}
\usepackage{amssymb,amsmath,amsfonts,soul,amsthm,enumerate}
\usepackage{graphicx}
\usepackage[cp1251]{inputenc}
\usepackage[T2A]{fontenc}
\usepackage{mathrsfs}
\usepackage{tabularx}
\usepackage[colorlinks,allcolors=blue]{hyperref} 
\usepackage[dvipsnames]{xcolor}
\usepackage[]{algorithm2e}
\usepackage{algpseudocode}
\usepackage{listings}

\def\@evenfoot{}
\def\@oddfoot{}

\begin{document}
\def\@evenhead{\vbox{\hbox to \textwidth{\thepage\leftmark}\strut\newline\hrule}}

\def\@oddhead{\raisebox{0pt}[\headheight][0pt]{%
\vbox{\hbox to \textwidth{\rightmark\thepage\strut}\hrule}}}

\newpage
\normalsize
\def\bibname{\vspace*{-30mm}{\centerline{\normalsize References}}}
\thispagestyle{empty}
\vskip 5 mm

\centerline{\bf COMPARATIVE ANALYSIS OF COLOR MODELS FOR HUMAN}
\centerline{\bf PERCEPTION AND VISUAL COLOR DIFFERENCE}

\vskip 0.3cm
\centerline{\bf Burambekova Aruzhan, Shamoi Pakizar}

\vskip 0.3cm
\noindent{\bf Abstract}
{\small 
Color is integral to human experience, influencing emotions, decisions, and perceptions. This paper presents a comparative analysis of various color models' alignment with human visual perception. The study evaluates color models such as RGB, HSV, HSL, XYZ, CIELAB, and CIELUV to assess their effectiveness in accurately representing how humans perceive color. We evaluate each model based on its ability to accurately reflect visual color differences and dominant palette extraction compatible with the human eye.   In image processing, accurate assessment of color difference is essential for applications ranging from digital design to quality control. Current color difference metrics do not always match how people see colors, causing issues in accurately judging subtle differences. Understanding how different color models align with human visual perception is crucial for various applications in image processing, digital media, and design. 

}
\vskip 0.3cm

\setcounter{figure}{0}

\renewcommand{\thesection}{\large 1}

\section{\large Introduction}
Color is one of the most influential object properties \cite{scis2023}. The way humans perceive colors is a complex and intriguing process, influenced by various physical, psychophysical, physiological, and psychological factors \cite{Cojocaru2024}. The perception of color is fundamentally subjective and varies from individual to individual \cite{muragul}. The brain plays a crucial role in decoding electrical signals into color experiences \cite{MoreiradaSilva2023}. The existence of 7.5 to 10 million colors and the role of color in cultural practices further highlight the complexity of color perception \cite{Nair2015InOT}. In computer graphics and digital imaging, accurately simulating this process is essential to create visually appealing and lifelike images.

The demand has driven the development of various color models \cite{9945709}, each with its own strengths and weaknesses in accurately representing human color perception.

This article provides a comparative evaluation of several prominent color systems, including RGB, CMYK, HSV, and CIE LAB\cite{9373145}, \cite{7557900}. We will explore their theoretical foundations and practical applications\cite{8574809}, as well as their effectiveness in capturing the nuances of human vision. Understanding these color systems can help us better appreciate the efforts to bridge the gap between digital representation and human perception.

Color difference, also known as color distance, is a measure that assigns how different colors are perceived by each other. Perceptually similar color pairs have smaller distances \cite{6561275}.

The search for an optimal color difference formula has spanned several decades, resulting in numerous options.
Existing formulas often incorporate specific attributes like lightness, chroma, and hue weighting functions to capture the complexity of color perception. However, despite extensive efforts, finding a universally effective formula that consistently reflects perceptual similarities and differences remains an ongoing problem.

\renewcommand{\thesection}{\large 2}
\section{\large Related Work}

The color difference is important in different perception-based image processing problems, for example, Lossy Image Compression, Color Gamut Mapping, Segmentation, and Image Enhancement \cite{Lissner2012}, \cite{6995129}. A range of studies have explored the concept of color distance, each proposing different models and methods for its measurement. 


The common way to calculate color distance is by using the CIE LAB color difference, denoted as $ \Delta E^*_{ab} $. This is calculated as the Euclidean distance between the two points representing the color stimuli in the space.\cite{CIE2004}


More advanced color difference formulas have been created, including CIE94 and CIEDE2000 \cite{Luo2001}. They aim to provide a better correlation with human perception.

Recent research proposed a novel model based on visual recognition \cite{Jingqin2018Color}. Another paper introduced a hybrid color distance model inspired by the human vision system \cite{Ramon2011Hybrid}.  \cite{Jaume2005Evaluation} evaluates the performance of various color distance measures in image segmentation, with the graph-partition approach and the Frechet distance yielding the best results. Next, \cite{C2010Color} proposes a luminance-invariant color distance based on parallel coordinates and \cite{D1998Distance} testing various vector distance measures for color image retrieval. Lastly,  \cite{W2004Color} presents a color difference model based on the receptors' properties in the honeybee's color vision system.


One common formulation, documented in CIEDE2000 \cite{Luo2001}, includes standard attributes such as lightness, chroma, and hue weighting functions. Additionally, it considers an interactive term between chroma and hue differences. This formulation has been shown to improve the performance of blue colors and includes a scaling factor for enhancing the performance of gray colors.


All color-difference formulas are designed to assess the color differences between pairs of stimuli separated by a fine line\cite{mirjalili2019color}. Based on the CIEDE2000 system, a novel color-difference formula has been developed for pairs of samples with no separation, encompassing a wide range of color differences smaller than 9.1 units on the CIEDE2000 scale. This new formula is designated the color-difference formula for the no separation viewing condition and is referred to as \(\Delta E_{NS}\).


Similar approaches CMC(I:c) and BFD(I:c) \cite{Luo1987} have better agreement with experimental results for small color differences between surface colors than other published formulas.


Most early researchers opted for Riemannian Color Spaces. However, recent color theory trends indicate a non-Riemannian nature of perceptual color space. For example, \cite{Bujack2022} shows that the principle of diminishing returns applies to human color perception.

Using self-luminous sample pairs on a CRT display, \cite{Cui2001} discovered that varying the separation size between the two samples had a minor impact on the perceived color difference. However, it did modify the weighting factor between lightness and chromatic differences.

\renewcommand{\thesection}{\large 3}
\section{\large Methods}
\subsection{Color Models}
For analysis, models like RGB, XYZ, and LAb were chosen.
\subsubsection{RGB Model}
The RGB color model is one of the most widely used methods for color representation in computer graphics. It employs a color coordinate system based on three primary colors: red, green, and blue. The combination of these colors in varying intensities forms the cube-shaped RGB color space, encompassing all colors that can be created through their linear combination.

\subsubsection{XYZ Model}

\begin{figure}[h]
    \centering
\includegraphics[scale=0.5]{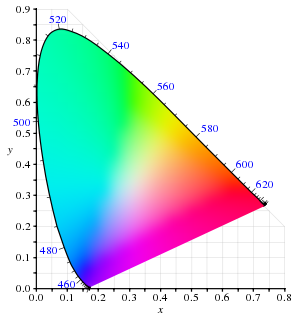}
\caption{The 1931 CIE chromaticity diagram (XYZ model) \cite{Hastings2012}.}
\label{xyz}
\end{figure}

The XYZ color model was created by the International Commission on Illumination (CIE) in 1931. It is based on human vision and uses three imaginary primary colors to map out all perceivable colors (see Fig. \ref{xyz}). The XYZ model forms the basis for many other color spaces and is used for color matching and comparison.

\subsubsection{LAB Model}
\begin{figure}[h]
    \centering
    \includegraphics[scale=0.2]{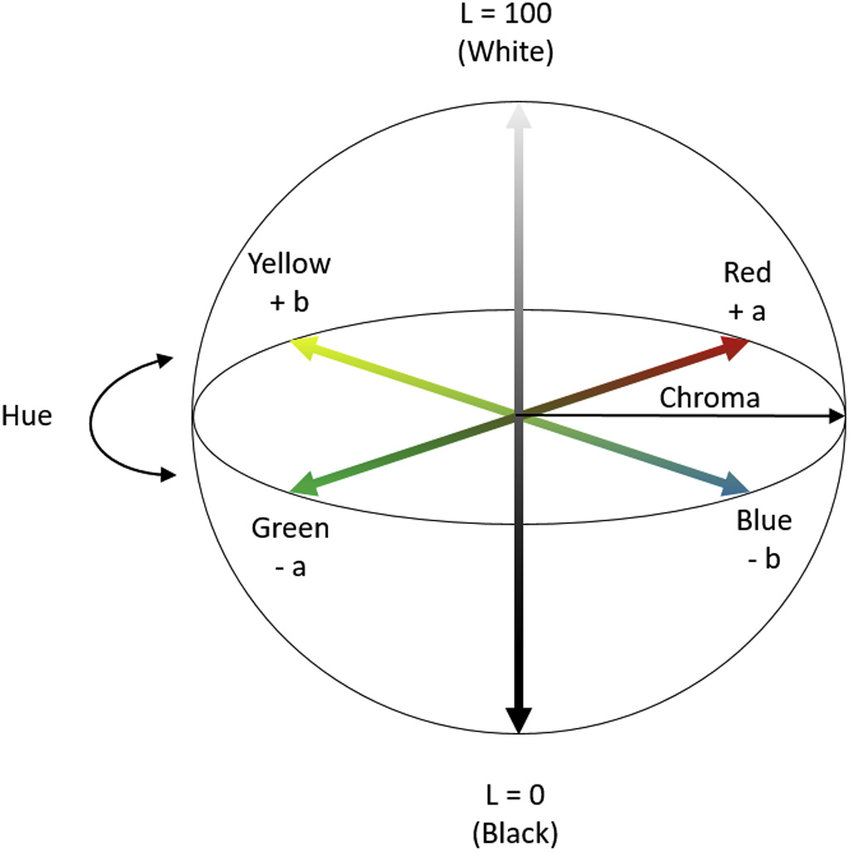}
    \caption{The CIELAB color space diagram \cite{imglab}}
    \label{cielab}
\end{figure}

    The LAB color model, or CIELAB, represents color in three dimensions: L* for lightness, a* for the green-red component, and b* for the blue-yellow component (see Fig. \ref{cielab}).
    It is designed to be perceptually uniform, meaning equal changes in value correspond to roughly equal changes in perceived color.
    LAB is widely used in industries where color accuracy is critical, such as printing and textile manufacturing.

The CIE 1976 (L\*a\*b\*) color difference formula is shown below in Eq. \ref{eq:cielab_difference}.
\begin{equation}
\Delta E_{ab}^* = \sqrt{(L_2^* - L_1^*)^2 + (a_2^* - a_1^*)^2 + (b_2^* - b_1^*)^2}
\label{eq:cielab_difference}
\end{equation}

The CIEDE2000 color difference formula is given by:

\begin{equation}
\Delta E_{00} = \sqrt{\left(\frac{\Delta L'}{k_L S_L}\right)^2 + \left(\frac{\Delta C'}{k_C S_C}\right)^2 + \left(\frac{\Delta H'}{k_H S_H}\right)^2 + R_T \left(\frac{\Delta C'}{k_C S_C}\right) \left(\frac{\Delta H'}{k_H S_H}\right)}
\label{eq:delta_e00}
\end{equation}

where the terms are defined as follows:

\begin{equation}
\Delta L' = L_2^* - L_1^*
\label{eq:delta_L}
\end{equation}

\begin{equation}
\Delta C' = C_2' - C_1'
\label{eq:delta_C}
\end{equation}

\begin{equation}
\Delta H' = 2 \sqrt{C_1' C_2'} \sin\left(\frac{\Delta h'}{2}\right)
\label{eq:delta_H}
\end{equation}

\begin{equation}
R_T = -2 \sqrt{\frac{C_7}{C_7 + 25^7}} \sin\left(2 \Delta \theta\right)
\label{eq:RT}
\end{equation}

\begin{equation}
S_L = 1 + \frac{0.015 (L_1^* - 50)^2}{\sqrt{20 + (L_1^* - 50)^2}}
\label{eq:SL}
\end{equation}

\begin{equation}
S_C = 1 + 0.045 C_1'
\label{eq:SC}
\end{equation}

\begin{equation}
S_H = 1 + 0.015 C_1' \left(1 - 0.17 \cos(\Delta h' - 30^\circ) + 0.24 \cos(2 \Delta h') + 0.32 \cos(3 \Delta h' + 6^\circ) - 0.20 \cos(4 \Delta h' - 63^\circ)\right)
\label{eq:SH}
\end{equation}

\subsubsection{CIELUV Model}
\begin{figure}[h]
    \centering
\includegraphics[scale=0.07]{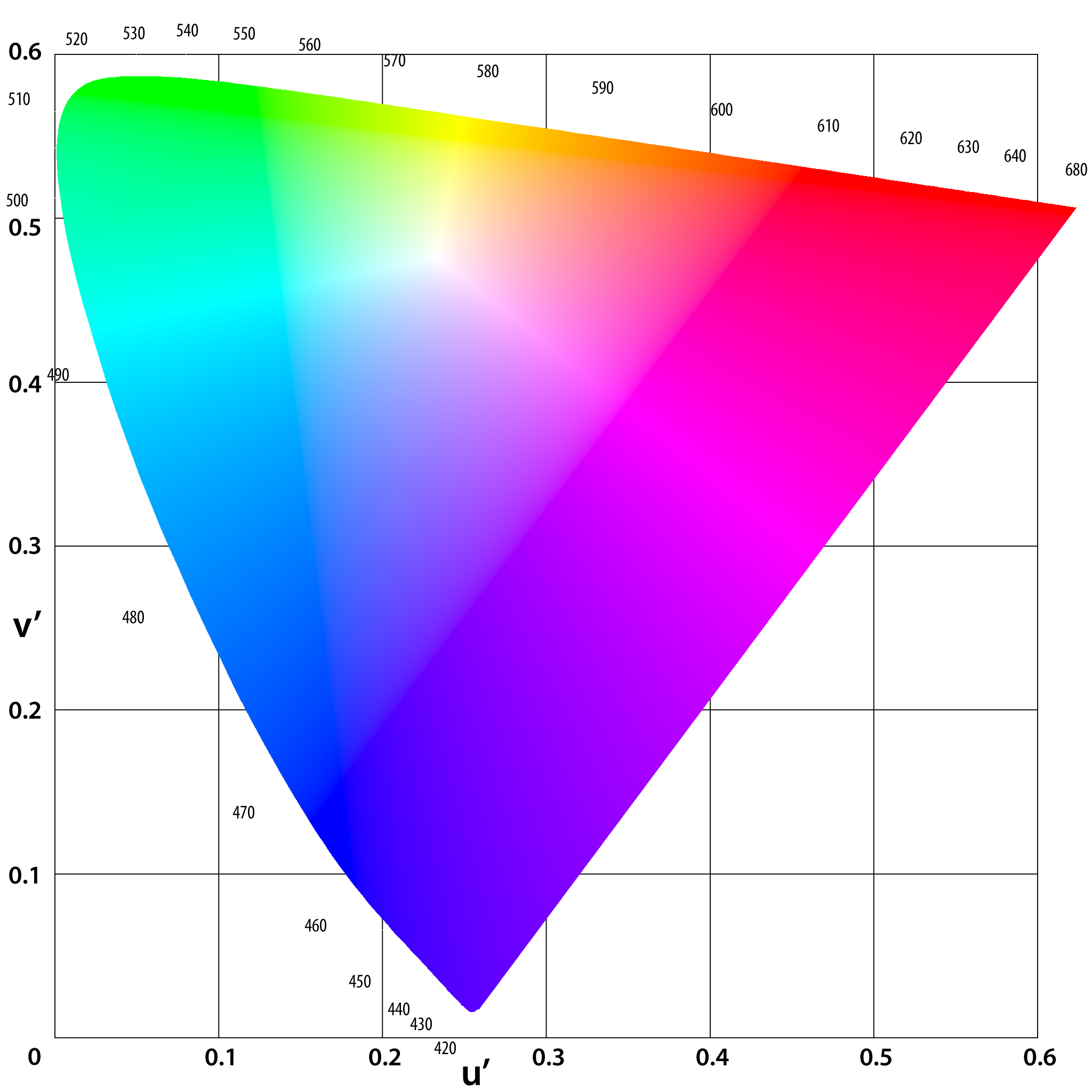}
\caption{CIELUV Model $(u', v')$ chromaticity diagram using the 1931 standard observer. (Photo downloaded from Wikimedia Commons: http://commons.wikimedia.org/.)}
\label{cieluv}
\end{figure}

    The LUV color model, or CIE LUV (see Fig. \ref{cieluv}), is similar to LAB but optimized for different applications.
     LUV is used primarily in computer graphics and digital imaging due to its linearity and ease of transformation to other color spaces.
    It provides a perceptually uniform color space for more accurate color representation.

\subsubsection{HSL Model}
    HSL stands for Hue, Saturation, and Lightness.
    It is often used in graphics software to select colors more intuitively.
    The HSL model represents colors in a cylindrical coordinate system, making it easier to independently adjust hue, saturation, and lightness.

\subsubsection{HSV Model}
\begin{figure}[h]
\centering
\includegraphics[scale=0.5]{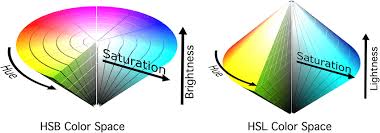}
\caption{HS* Models.  (Photo downloaded from Wikimedia Commons: http://commons.wikimedia.org/.)}
\label{hs}
\end{figure}

 HSV(HSB) stands for Hue, Saturation, and Value.
    It is similar to HSL but uses value instead of lightness, which can make it more intuitive for certain applications (see Fig. \ref{hs}).
    The HSV model is commonly used in image editing software and for color-picking tools.

Basically, conversion from each model are widely used in image processing. For example, the conversion from XYZ to CIELAB is presented in Eq. \ref{eq:cielab_conversion}
\begin{equation}
\begin{aligned}
L^* &= 116 \cdot f\left(\frac{Y}{Y_n}\right) - 16 \\
a^* &= 500 \cdot \left[f\left(\frac{X}{X_n}\right) - f\left(\frac{Y}{Y_n}\right)\right] \\
b^* &= 200 \cdot \left[f\left(\frac{Y}{Y_n}\right) - f\left(\frac{Z}{Z_n}\right)\right]
\end{aligned}
\label{eq:cielab_conversion}
\end{equation}
where
\begin{equation}
f(t) = \begin{cases}
t^{1/3} & \text{if } t > \left(\frac{6}{29}\right)^3 \\
\frac{1}{3}\left(\frac{29}{6}\right)^2 t + \frac{4}{29} & \text{otherwise}
\end{cases}
\label{eq:cielab_f}
\end{equation}

\subsection{k-means Clustering}

We employ the k-means clustering algorithm \cite{kmeans1} \cite{kmeans2} in various color models to extract dominant colors from an image. The process involves the following steps:
\begin{itemize}
    \item    Convert the image from its original color space (typically RGB) to the desired color model (e.g., HSV, HSL, CIELAB).
    \item Reshape the image data into a 2D array where each row represents a pixel, and each column represents a color channel in the selected color model.
    \item Apply the k-means clustering algorithm to the 2D array to find \( k \) clusters representing the dominant colors. The objective function of k-means clustering is to minimize the sum of squared distances between each pixel and the centroid of its assigned cluster:
   \[
   \min \sum_{i=1}^{k} \sum_{\mathbf{x} \in C_i} \|\mathbf{x} - \mathbf{\mu}_i\|^2
   \]
   where \( \mathbf{x} \) is a pixel, \( C_i \) is the set of pixels assigned to cluster \( i \), and \( \mathbf{\mu}_i \) is the centroid of cluster \( i \).
   \item The centroids of the \( k \) clusters represent the dominant colors in the image.
   \[
   \mathbf{\mu}_i = \frac{1}{|C_i|} \sum_{\mathbf{x} \in C_i} \mathbf{x}
   \]
   where \( \mathbf{\mu}_i \) is the centroid of cluster \( i \), and \( |C_i| \) is the number of pixels in cluster \( i \).
   \item Convert Dominant Colors Back to RGB for display and further analysis.
\end{itemize}
By applying these steps, we can extract the dominant colors from an image using different color models, allowing us to analyze and compare their effectiveness in representing the image's color characteristics.

\subsection{Survey design}

\begin{figure}[h]
    \centering
    \includegraphics[width=0.4\linewidth]{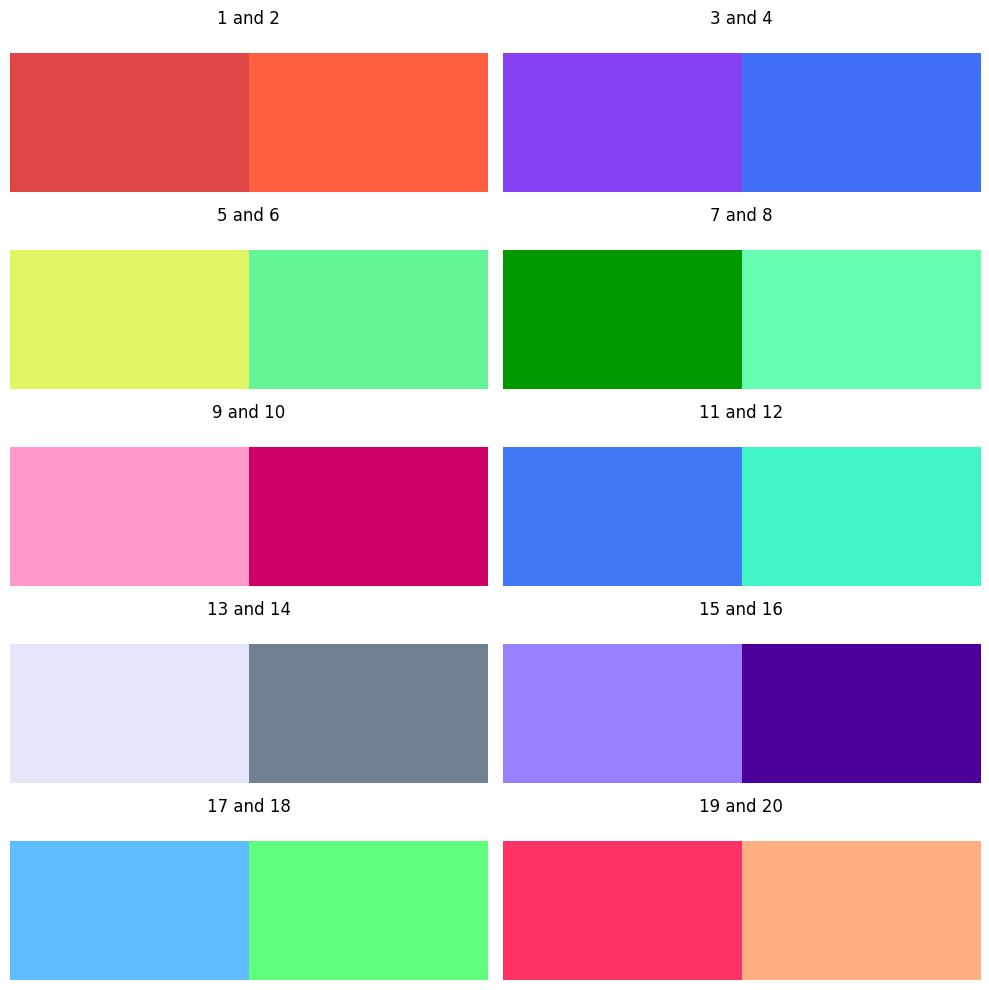}
    \caption{Color Pairs used in the survey}
    \label{fig:pairs}
\end{figure}

In our experiments, 15 human participants (7 males and 8 females aged 20 to 23 years old) took part in a survey to evaluate perceived color differences between color pairs. The experiment aimed to understand how well various color models and metrics align with human visual perception. A set of color pairs was presented to each participant (see Fig. \ref{fig:pairs}). The color pairs were selected to cover a range of hues, saturations, and brightness levels. The perceived differences provided by the participants were collected and averaged for each color pair.

\renewcommand{\thesection}{\large 4}
\section{\large Results}

\begin{figure*}[h]
    \centering
    \includegraphics[width=\textwidth]{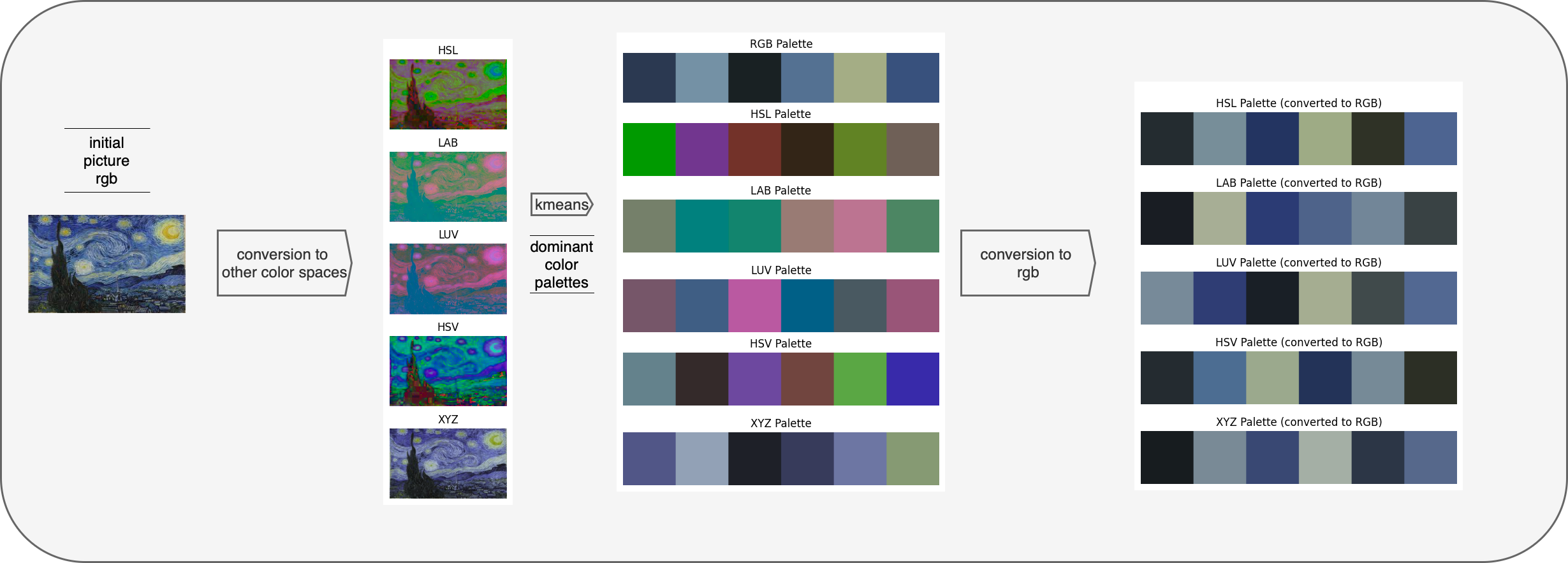}
    \caption{K-means based dominant palettes extraction: comparison between color models}
    \label{fig:palletes}
\end{figure*}

Two experiments were formulated to determine which of the most popular color spaces is closer to human perception. The first experiment (see Fig.\ref{fig:palletes}) involves obtaining dominant color palettes from an image using k-means\cite{9869653}, \cite{9163111} in different color models. The second experiment compares the results obtained from calculating the color differences between sets of color pairs using different color models and survey results from people about how they perceive these pairs of colors.

The first experiment employed Vincent van Gogh's famous painting, "The Starry Night." The celestial body depicted prominently in the sky is the bright yellow moon. The experiment showed that the colors in the picture are not all the same as those that a person would see when looking at it. Among the considered models, HSL resulted in the most human-consistent color palette.

\begin{figure*}[h]
    \centering
    \includegraphics[width=0.6\linewidth]{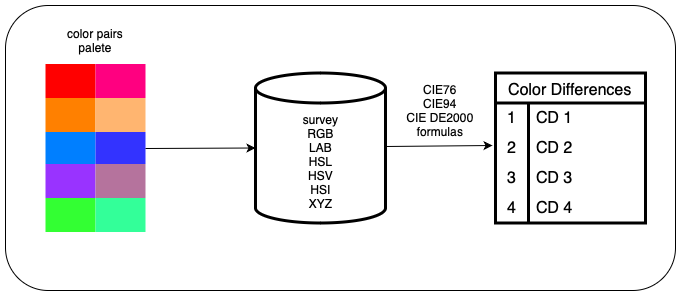}
    \caption{Color Differences}
    \label{fig:color differences}
\end{figure*}

The second experiment analysis involves estimating color differences between two colors using Euclidean Distance, Weighted Euclidean Distance, CIE 2000, CIE CMC(1:c), and distance in a cylindrical coordinate system(see Fig. \ref{fig:color differences}).

Experimental data is presented in Table \ref{analysis}.


HSL showed the highest correlation (0.72) with human perception, indicating a strong positive relationship.
HSV also demonstrated a significant positive correlation (0.60), suggesting it aligns well with human perception.

\begin{table*}[]
\centering
\caption{Results of the experiment on color difference across various color models}

\begin{tabularx}{\linewidth}{|X|X|X|X|X|X|X|X|X|X|X|}
\hline
Color Pair & Euclid. RGB & w\_RGB & LAB CIE2000 & LAB CMC & HSV    & HSL    & XYZ    & CMC CIE LUV & Human \\ \hline
1, 2    & 39.37         & 72.08         & 10.42       & 11.15   & 0.1304 & 0.2833 & 14.95  & 7.32        & 6     \\ \hline
3, 4    & 83.76         & 142.05        & 15.94       & 14.58   & 0.0003 & 0.0003 & 5.44   & 11.47       & 3     \\ \hline
5, 6    & 134.63        & 216.96        & 20.01       & 22.65   & 0.0006 & 0.0006 & 30.01  & 25.58       & 4     \\ \hline
7, 8    & 229.11        & 391.30        & 29.01       & 33.62   & 0.5657 & 0.4000 & 83.69  & 18.81       & 5     \\ \hline
9, 10   & 190.82        & 350.73        & 29.23       & 27.03   & 0.6325 & 0.4000 & 70.30  & 24.32       & 5     \\ \hline
11, 12  & 132.85        & 260.87        & 51.08       & 69.14   & 0.0004 & 0.0004 & 58.14  & 53.40       & 3     \\ \hline
13, 14  & 188.58        & 323.92        & 30.41       & 28.26   & 0.4393 & 0.6966 & 110.57 & 15.30       & 5     \\ \hline
15, 16  & 180.88        & 326.44        & 35.27       & 32.58   & 0.6418 & 0.4510 & 78.51  & 17.62       & 6     \\ \hline
17, 18  & 145.34        & 247.51        & 52.25       & 68.85   & 0.0006 & 0.0006 & 75.59  & 65.13       & 2     \\ \hline
19, 20  & 128.55        & 253.57        & 30.10       & 29.83   & 0.3137 & 0.1569 & 35.91  & 32.29       & 2     \\ \hline
\end{tabularx}
\label{analysis}
\end{table*}

\begin{figure}[h]
    \centering
    \includegraphics[width=0.65\linewidth]{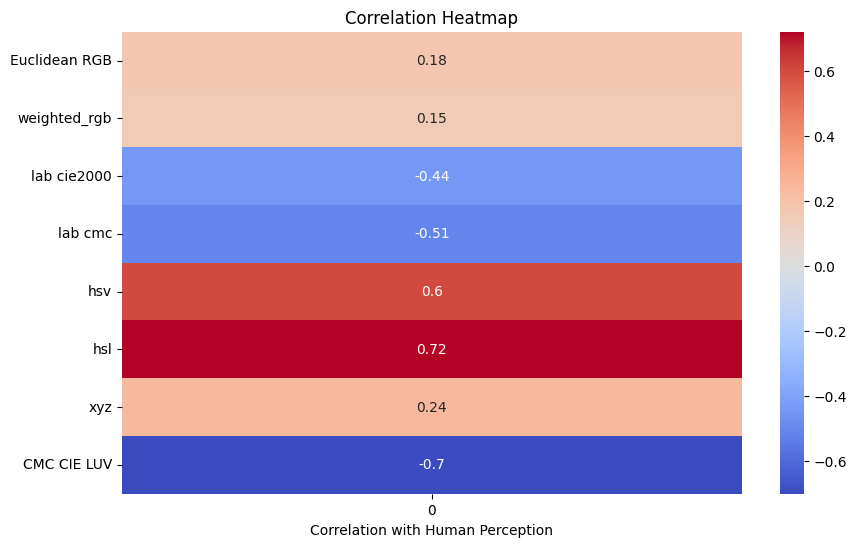}
    \caption{Heatmap illustrating the correlations between estimated color differences of various color models and human perception}
    \label{fig:heatmap}
\end{figure}

Moderate positive correlation (0.24) of XYZ indicates some alignment.
Euclidean RGB and weighted RGB  got low positive correlations (0.18 and 0.15, respectively), suggesting a limited relationship.
However, some formulas on models got negative correlations: Lab cie2000, Lab CMC, and CMC CIE LUV.(\ref{fig:heatmap})




\renewcommand{\thesection}{\large 5}
\section{\large Conclusion}
Existing color difference measures may not accurately align with human perception, leading to inconsistency in determining color variations. It is challenging to define color closeness due to non-uniformity and high correlation among their components in different spaces. The current paper presents our preliminary results on comparing different color models and how they relate to human vision. It analyzes the effectiveness of different color systems, such as RGB, HSV, HSL, and others, to see how well they represent how we see color.

This is important in image processing because it's essential to accurately judge color differences in digital design and quality control. However, current methods for measuring color differences don't always match up with how we actually see colors. 

Along with different color spaces, there are also different formulas for calculating color differences, like CIE 1976 and CMC(1:c).
HSL and HSV have the highest correlation and lowest MAE, making them the best models for aligning with human color perception. 
CMC CIE LUV has a pretty average MAE, but there's a negative correlation, which has mixed results. The Euclidean RGB, weighted RGB, Lab CIE 2000, and Lab CMC all have low correlation and a high MAE. They showed themselves to be the worst in comparison with human perception.

The study is limited by a small sample size, which may not reflect real-world conditions. Future work could expand the participant pool and explore additional color models to improve alignment with human perception. 

\section*{\large Acknowledgement}
This research has been funded by the Science Committee of the Ministry of Science and Higher Education of the Republic of Kazakhstan (Grant No. AP22786412)
\bibliography{bibfile}

\begin{flushleft}
Burambekova Aruzhan,\\
School of Information Technology and Engineering,
Kazakh-British Technical University, \\
Tole Bi street 59, Almaty, Kazakhstan,
Email: {\tt a\_burambekova@kbtu.kz},\\

Shamoi Pakizar,\\
School of Information Technology and Engineering,
Kazakh-British Technical University, \\
Tole Bi street 59, Almaty, Kazakhstan,
Email: {\tt p.shamoi@kbtu.kz},\\
...\\
\end{flushleft}

\end{document}